# Innovative Texture Database Collecting Approach and Feature Extraction Method based on Combination of Gray Tone Difference Matrixes, Local Binary Patterns, and K-means Clustering

Shervan Fekri-Ershad
Department of Computer Science and Engineering
Shiraz University
Shiraz, Fars, Iran
shfekri@shirazu.ac.ir

*Abstract-* Texture analysis and classification are some of the problems which have been paid much attention by image processing scientists since late 80s. If texture analysis is done accurately, it can be used in many cases such as object tracking, visual pattern recognition, and face recognition. Since now, so many methods are offered to solve this problem. Against their technical differences, all of them used same popular databases to evaluate their performance such as Brodatz or Outex, which may be made their performance biased on these databases. In this paper, an approach is proposed to collect more efficient databases of texture images. The proposed approach is included two stages. The first one is developing feature representation based on gray tone difference matrixes and local binary patterns features and the next one is consisted an innovative algorithm which is based on K-means clustering to collect images based on evaluated features. In order to evaluate the performance of the proposed approach, a texture database is collected and fisher rate is computed for collected one and well known databases. Also, texture classification is evaluated based on offered feature extraction and the accuracy is compared by some state of the art texture classification methods.

## I. INTRODUCTION

Texture is a description of the spatial arrangement of color or intensities in an image or a selected region of an image. There are some aspects about textures such as size of granularity, directionality, randomness or regularity and texture elements. Analysis these aspects may be useful to classify the textures in one or more classes. For example, four different textures of grass, leaves, Wood, and Brick wall are shown in Fig.1, which are different in many terms such as regularity, directionality and etc.

Texture analysis and classification are some of the problems which have been paid much attention by image processing scientists since late 80s. Texture analysis is one of the basic stages in computer vision cases such as object tracking, visual pattern recognition, face recognition, skin detection, image retrieval and etc.

Since now, many approaches were offered to solve this problem accurately. These approaches are difference in technical details. Some of them directly work on the taken images of the textures, such as texture classification based on random threshold vector technique [1], and texture classification based on primitive pattern units [2]. Another group of methods, first do some process on the images and then search for suitable features related to the class labels, such as texture classification by using advanced local binary patterns, and spatial distribution of dominant patterns [3] and new color texture approach for industrial products inspection [4]. Xie. categorized the techniques used to texture analysis in four categories, statistical approaches, Structural approaches filter based methods, and model based approaches [5].

Some approaches proposed a robust algorithm which is combination of structural and statistical features. For example, in [6] an approach is proposed for classification of medical Indian plants based on combination of color, texture and edge features together.

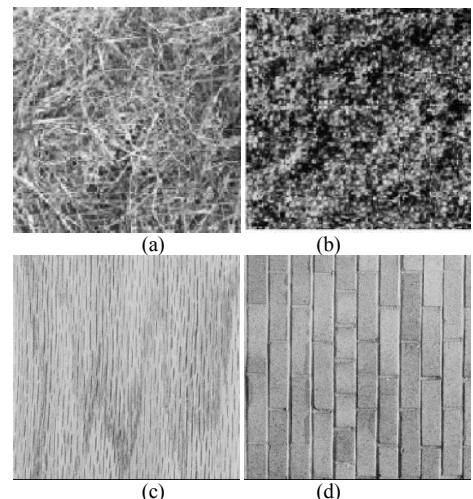

Figure.1. Some examples of texture
(a) Grass (b) Leaves (c) Wood (d) Brick Wall

Table 1 shows a summary list of some of the key texture analysis methods that have been applied to texture classification, segmentation or analysis. Methods are grouped in 4 categories. Clearly, statistical and filter based approaches have been more popular than structural and model-based approaches. In [5], fundamental of these methods are discussed with details.

Table 1: Inexhaustive list of textural analysis methods

| Category | Method |
|---|---|
| Statistical | 1. Histogram properties<br>2. Co-occurrence matrix<br>3. Local binary pattern<br>4. Other gray level statistics<br>5. Autocorrelation<br>6. Registration-based |
| Structural | 1. Primitive measurement<br>2. Edge Features<br>3. Skeleton representation<br>4. Morphological operations |
| Filter Based | 1. Spatial domain filtering<br>2. Frequency domain analysis<br>3. Joint spatial/spatial-frequency |
| Model Based | 1. Fractal models<br>2. Random field model<br>3. Texem model |

Regardless, technical differences between proposed texture analysis techniques, researchers evaluate their approach's performance, on same popular databases such as Brodatz[7] and Outex[8]. This similarity may make them biased on test databases.

In this respect, there are some basic questions here that, why these databases are used? How much are these natural? Or, how much are different their images in terms of regularity, directionality or etc? Briefly, why these texture images are collected in a database?

Since late 90s, nearly all of the proposed texture classification methods considered Outex and Brodatz databases as the train and test images. It may make methods and related performance biased to these database images. This problem absolutely decrease the accurately rate versus untrained images in real applications. Some researchers were proposed texture analysis techniques with high accuracy rate which are not useful in special applications because of biasing to the popular test databases.

In this paper, some features are proposed to analyze and classify textures which are more discriminative than previous. In order to have a more general database which includes high range of natural textures, an algorithm is proposed to collect images using combination of proposed feature representation and k-means clustering algorithm.

## II. FEARTURE EXTRACTION TO ANALYSIS TEXTURES

The aim of this section is to extract some features that discriminate textures. In this respect, literature survey of feature extraction is studied to collect discriminative ones together as a combination vector. Since now, many features are proposed to analysis textures such as Local binary patterns (LBP)[9], Haralick features[10], Voronoi polygon features[11], histogram properties[12], Gray level co-occurrence matrixes(GLCM) [13], Primitive pattern units[14], and etc.

There is a large overlap between these features. For example, Haralick features such as entropy and energy can be computed based on histogram properties. My researches are shown that some features are more discriminative when considered together as a combined feature vector. Also, these features are more efficient to analyze natural textures.

*A. Gray Tone Difference Matrixes*

A Grey-Tone Difference Matrix (GTDM) was suggested in [15] in an attempt to define texture measures correlated with human perception of textures. A GTD matrix is a column vector containing *G* elements. Its entries are computed based on measuring the difference between the intensity level of a pixel and the average intensity computed over a square, sliding window centered at the pixel. Suppose the image intensity level *f(x,y)* at location *(x,y)* is *i*, *i=0,1,...,G-1*. The average intensity over a window centered at *(x,y)* is

$$\overline{f_i} = \overline{f}(x,y) = \frac{1}{W-1} \sum_{m=-k}^{k} \sum_{n=-k}^{k} f(x+m, y+n)$$

Where K specifies the window size and *W = 2(2K+1)*. The $i_{th}$ entry of the gray-tone difference matrix is:

$$S_i = \sum_{x=0}^{M-1} \sum_{y=0}^{N-1} |i - \overline{f_i}| \qquad (2)$$

Notice, for all pixels having the intensity level *i* and otherwise, *s(i) = 0*.

Three different features were derived from the GTDM, to quantitatively describe such perceptual texture properties as follows:

- Coarseness (defined by the size of texture primitives):

$$f_1 = \left(\varepsilon + \sum_{i=0}^{G-1} P_i s(i)\right)^{-1} \qquad (3)$$

Where, ε is a small number to prevent the coarseness coefficient becoming infinite and $p_i$ is the estimated probability of the occurrence of the intensity level *i* and *pi = Ni / n*. Also, with *Ni* denoting the number of pixels that have the level *i*, and n = (N-K)(M-K).

- B. complexity (dependent on the number of different primitives and different average intensities):

$$f_2 = \sum_{i=0}^{G-1} \sum_{j=0}^{G-1} \frac{|i-1|}{n(p_i+p_j)} [p_i s(i) + p_j s(j)], \ p_i, p_j \neq 0 \quad (4)$$

- C. texture strength (clearly definable and visible primitives):

$$f_3 = \frac{\sum_{i=0}^{G-1}\sum_{j=0}^{G-1}(p_i+p_j)(i-j)^2}{\epsilon+\sum_{i=0}^{G-1} s(i)} \quad (5)$$

Some other features such as contrast and busyness can be derived using GTDM, but the result shows discrimination effect of these features are not much in this case.

*B. Histogram Features*

The histogram can be easily computed, given the image. The shape of the histogram provides many clues as to the character of the texture. For example, a narrowly distributed histogram indicated the low-contrast texture. A bimodal histogram often suggests that the texture contained an object with a narrow intensity range against a background of differing intensity. Different useful parameters (image features) can be worked out from the histogram to quantitatively describe the first-order statistical properties of the image.

Most often the so-called central moments are derived from it to characterize the texture. Among them, mean, skewness, kurtosis, energy and entropy are more reliable and discriminant for texture analysis and classification. These are defined by (6) to (10) as below.

Mean: $f_4 = \sum_{i=0}^{G-1} i P(i)$ (6)

Skewness: $f_5 = \sigma^{-3} \sum_{i=0}^{G-1}(i-f_1)^3 p(i)$ (7)

Kurtosis= $f_6 = \sigma^{-4} \sum_{i=0}^{G-1}(i-f_1)^4 p(i) - 3$ (8)

Energy = $f_7 = \sum_{i=0}^{G-1}[p(i)]^2$ (9)

Entropy= $f_8 = \sum_{i=0}^{G-1} p(i) \log_2[p(i)]$ (10)

Where, G is total number of intensity levels in image, *p(i)* means the occurrence probability of $i_{th}$ level of intensity in image. σ is the total variance of the image.

The mean takes the average level of intensity of the image or texture being examined. The skewness is zero if the histogram is symmetrical about the mean, and is otherwise either positive or negative depending whether it has been skewed above or below the mean. The kurtosis is a measure of flatness of the histogram; the component '3' inserted in (8) normalizes $f_4$ to zero for a Gaussian-shaped histogram. The entropy is a measure of histogram uniformity. Other possible features derived from the histogram are the minimum, the maximum, the range and the median value.

*C. Local Binary Patterns*

The local binary patterns (LBP) is a non-parametric operator which describes the local spatial structure and local contrast of an image. Ojala et al. [16] first introduced this operator and showed its high discriminative power for texture classification.

At a given pixel position *(x_c, y_c)*, LBP is defined as an ordered set of binary comparisons of pixel intensities between the center pixel and its surrounding pixels. Usually to achieve the rotation invariant, neighborhoods would be assumed circular. So, points which the coordination's are not exactly located at the center of pixel would be found by interpolation.

Now, the local binary patterns are defined at a neighborhood of image by (11).

$$LBP_{P,R} = \sum_{n=0}^{p-1} s(g_n - g_c) 2^n \quad (11)$$

Where, "$g_c$" corresponds to the grey value of the centered pixel and "$g_n$" to the grey values of the neighborhood pixels. Also, P is the number of neighborhoods of center pixel, and function s(x) is defined as:

$$s(x) = \begin{cases} 1 & \text{if } x \geq 0 \\ 0 & \text{if } x < 0 \end{cases} \quad (12)$$

According to [25], The $LBP_{P,R}$ operator produces ($2^P$) different output values, corresponding to the $2^P$ different binary patterns that can be formed by the P pixels in the neighbor set.

The author's practical experience in [17], showed that computation complexity of basic local binary patterns is too high. To solve this, Ojala et al [9] defined an uniformity measure "U", which corresponds to the number of spatial transitions (bitwise 0/1 changes) in the "pattern". It is shown in (13). For example, patterns 00000000 and 11111111 have U value of 0, while 00011101 have U value of 3.

$$U(LBP_{P,R}) = |s(g_{P-1} - g_c) - s(g_0 - g_c)| + \sum_{i=1}^{p-1}|s(g_i - g_c) - s(g_{i-1} - g_c)| \quad (13)$$

In this version of LBP, the patterns which have uniformity amount less than $U_T$ are categorized as uniform patterns and the patterns with uniformity amount more than $U_T$ categorized as non-uniform patterns. Finally, the LBP is computed by using (14).

$$LBP_{P,R}^{riu_T} = \begin{cases} \sum_{i=0}^{P-1} s(g_i - g_c) & \text{if } U(LBP_{P,R}) \leq U_T \\ P+1 & \text{otherwise} \end{cases} \quad (14)$$

Superscript "$riu_T$" reflects the use of rotation invariant "uniform" patterns that have U value of at most $U_T$.

According to (14), applying LBP will assign a label from 0 to *P* to uniform patterns and label *P+1* to non-uniform patterns.

Because, in this version of LBP just one label (P+1) is assigned to all of the non-uniform patterns, so uniform labels should cover mostly patterns in the image. In [18], authors show that if in the definition of LBP operator the value of $U_T$ is selected equal to (P/4), only a negligible portion of the patterns in the texture takes label P+1.

The LBP operator can describe and use in a similar way for every size of neighborhoods like 5×5 or 7×7 or etc. Using 3×3 size for LBP operator, 10 uniform and non-uniform labels are grouped. So, the feature extraction can described based on this labels. In order to determine the feature vector, first should compute the probability of occurrence of each label in image as follows:

$$\text{Probability of Label K} = \frac{\text{Number of Pixels (LBP = K)}}{M \times N} \quad (15)$$

Where, M and N are the size of image. Each dimension of feature vector is one of the labels and that value is the occurrence probability of that label. For example, by using 3×3 LBP operator, a feature vector can provide which has 10 dimensions. In a similar way, for every LBP operators, feature vector can compute. For example, by using size 5×5 or 7×7 for LBP, the feature vectors are computed by 16 or 24 dimensions.

*D. Proposed Discriminant Feature Vector for Texture Analysis*

In order to collect a texture database, the first stage is providing a feature representation for each input candidate image. According to the previous sections, some metrics are described which can be derived for each image. In this respect, the feature vector $F$ can be described as follows, which can be a good identifier for each input candidate.

$$F_i = \langle f_1, f_2, \ldots, f_8, f_9, \ldots, f_{18} \rangle$$

Where, $F_i$ means the feature vector which is computed for $i_{th}$ candidate input image. Also, $f_1, f_2, .. f_8$, are the feature values which can be computed using (3) to (10). The dimensions number 9 to 18 are related to the features extracted using (15) which are based on local binary patterns.

Any sub set of 18 features may be used to evaluate the proposed approach, but according to our research feature numbers 9 to 18 are shown a specific class of pixels in image[18]. In this respect, decreasing some dimension may decrease the accuracy.

### III. PROPOSED ALGORITHM TO COLLECT TEXTURE IMAGE DATABASE

The aim of this section is to propose an algorithm to collect a texture images database. In order to collect a database in artificial intelligence cases, some motivations should be considered as follows:

- The database should covered all search space which is really may be occurred in the case
- The search space should be balanced in terms of samples and its feature values

Suppose, there many different texture images as candidates to join desired database. In the previous section, an identifier feature vector is presented to analysis each texture candidates. Each texture candidates is a sample which can be joined to the database or not. In order to collect a database which handled two up motivations, an algorithm is proposed in this section which is based on K-Means Clustering as follows:

1. Let "*N*", as the total number of images in desired database.
2. Compute the proposed feature vector F for all of the texture candidates as it is proposed in (16). The computed feature vector for $i_{th}$ candidate is notified by $F_i$.
3. Normalize the range dimension values of extracted vectors.
4. Choose *N* number of candidates randomly as cluster centers
5. Compare the distance between each one of the Candidate images and all centers using Soergel distance [22]. In order to choose the best distance metric, many other are studied [23], finally, Soergel is provided most accuracy. It is shown in (17)

$$D_{i_{th}-cj} = \frac{\sum_{k=1}^{18} |f_{ik} - f_{jk}|}{\sum_{k=1}^{18} \max(f_{ik}, f_{jk})} \quad (17)$$

Where, $Di_{th-Cj}$ is shown the distance between $i_{th}$ candidate and desired $j_{th}$ cluster center. Also, $K$ is shown the number of features and $f_{ik}$ is shown the value of $k_{th}$ normalized dimension in $F_i$.

6. Assign each candidate to nearest center and make *N* clusters
7. Compute the average feature vector for each cluster, using (18). It is notified by $F_{avg-Ci}$. Next, choose new computed $F_{avg-Ci}$ as updated cluster centers.

$$F_{avg-Ci} = \frac{1}{C_i N} \sum_{j=1}^{C_i N} \sum_{k=1}^{18} f_{jk} \quad (18)$$

Where, $C_iN$ means the total members of $i_{th}$ cluster.
8. Repeat steps 5 to 7 and update cluster centers.
9. Terminate the algorithms when cluster centers didn't change in two following iterations
10. Select nearest candidate textures to each cluster center as a member of desired dataset

In Fig.2 a block diagram of proposed texture image database algorithm is shown.

### IV. RESULTS

The main aim of this paper is discussed in two categories. The first goal is to propose some features to analysis textures and the second one is proposing a texture collection algorithm. In order to evaluate the performance of the first aim, texture classification based on proposed feature vector may be a good idea. In this respect, one of the popular texture databases which is called Brodatz is considered.

Brodatz dataset is included 112 classes of textures. In this dataset, each class is included one texture image in size of 512×512. In order to provide unbiased test and train set, 12 classes are chosen randomly. In order to provide satisfied number of train and test images for classifying, each one is cropped to windows in size of 128×128 and overlapped by size 64. In this respect, 48 images are provided for each class. At random, 40% of images are considered as train and 60% for test set [19]. The proposed feature extraction approach is applied on them. Finally, texture classification is done using K-nearest neighborhood as classifier. This process is repeated 10 times and the total classification rate is shown in the table 2. Also, J48 tree and Bayesian network are used as classifier in a same way. Notice, the computed feature vectors are normalized in each dimension based on the dimension range.

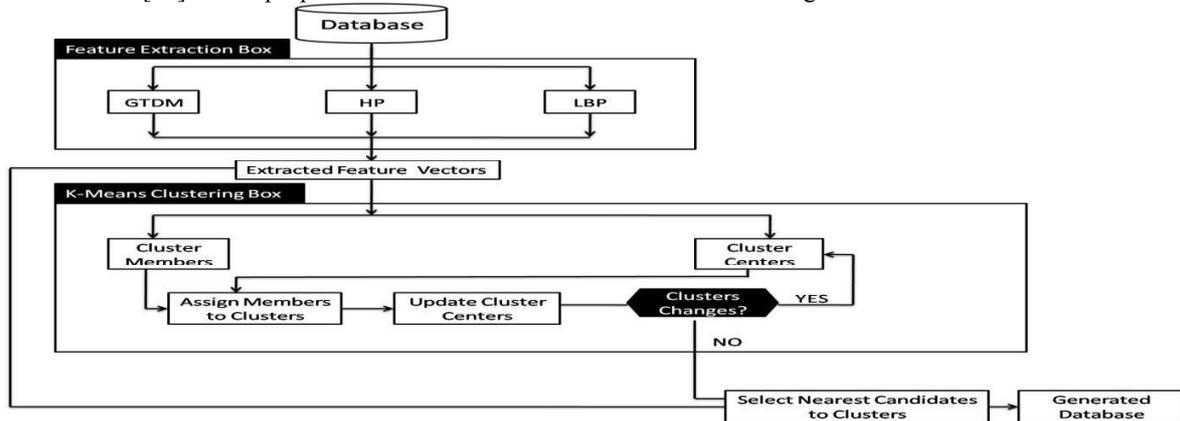

Figure 2. The Block Diagram of the proposed database collecting algorithm

In order to compare the performance of the proposed feature extraction algorithm and other methods, classification is done using some other well-known, efficient and new feature representation methods such as combination of primitive pattern units and statistical features [2], random threshold vector [1], and energy variation [20]. The classification rates are shown in table2. Some other classifiers such as J48 tree and Bayesian Network are used. Respected, results are shown in table3.

Table2. The classification rate of proposed approach and some other methods using KNN classifier

| 1NN | | | | |
|---|---|---|---|---|
| Method | Proposed | [Ers11] | [R+10] | [Ers12] |
| Accuracy rate | **_98.97±0.1_** | 93.31±0.5 | 88.17±0.1 | 89.46±0.2 |
| 3NN | | | | |
| Method | Proposed | [Ers11] | [R+10] | [Ers12] |
| Accuracy rate | **_99.14±0.2_** | 95.23±0.5 | 91.11±0.4 | 94.20±0.6 |
| 5NN | | | | |
| Method | Proposed | [Ers11] | [R+10] | [Ers12] |
| Accuracy rate | **_98.63±0.3_** | 94.31±0.4 | 90.62±0.2 | 90.07±0.6 |

Table3. The classification rate of proposed approach and some other methods using J48 and B.N. Classifiers

| J48 Tree | | | | |
|---|---|---|---|---|
| Method | Proposed | [Ers11] | [R+10] | [Ers12] |
| Accuracy Rate | **_93.70±0.5_** | 92.65±0.2 | 84.47±0.5 | 91.1±0.4 |
| Bayesian Network | | | | |
| Method | Proposed | [Ers11] | [R+10] | [Ers12] |
| Accuracy Rate | **_94.89±0.3_** | 94.20±0.3 | 85.49±0.4 | 93±0.1 |

As, it is shown in the tables2 and 3, the proposed feature representation has satisfied quality to classify textures accurately. The best accuracy rate is provided by proposed approach based on 3NN as classifier.

As it is seen in table3, the proposed approach is more efficient and accurate than other ones. Exactly, 94.89 rates, which is provided by proposed approach using Bayesian network, is the best in table 3.

In order to evaluate performance of the proposed approach for collecting database, a novel test is proposed as follows:

First of all, some texture images are captured by a digital camera resolution 200dpi. Captured textures are included grass, stone, wall, fences and etc. Next, N texture images are collected based on proposed algorithm in section 3. The proposed feature representation is applied on them, and N feature vectors are provided. This process is applied exactly on popular databases such as Brodatz and Outex. Notice, the number of databases is considered equal to N.

According to the mentions that are discussed in the section3, therefore, we will be looking for a projection where examples from the same class are projected very close to each other and, at the same time, the projected means are as farther apart as possible. This discussing is shown in Fig.2. In [21], the fisher measure is discussed to show the quality of clustering approaches. In our case, each one of the chosen images have a role same as a cluster center. According to the numerator of fisher equation, how much the cluster centers are far from others, clustering algorithm is covered more search space. It is shown in (19).

$$\text{Fisher} = 1/C^2 \sum_{i=1}^{C} \sum_{j=1}^{C} |D(\mu_i - \mu_j)| \qquad (19)$$

Where, $\mu_i$ is the mean of $i_{th}$ cluster center which is equal by $i_{th}$ member in our collected dataset. $C$ is the number of classes, which is equal by N in our collected dataset. Also, $1/C^2$ is used to normalize the computed fisher value based on size of datasets.

In the table4, the numerator of fisher measure is computed for two example clusters.

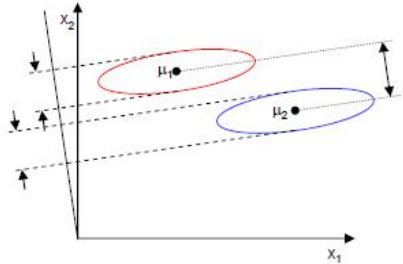

Figure.3. The desired projection of collected images

In the Fig. 3, two clusters are shown as example. Each cluster centers are notified by $\mu_1$ and $\mu_2$. In this dataset, the members are too far from each other; hence, the dataset covered a vast search space. The fisher measure may be high for this dataset.

Table4. The fisher value of proposed collected database and some popular texture dataset

| Dataset | Proposed Collected | Brodatz | Outex | Simplicity |
|---------|-------------------|---------|-------|------------|
| Fisher  | _0.693_           | 0.472   | 0.510 | 0.418      |

The results are shown that the collected database is covered more spaces of the texture analysis search space. Respectively, as it is seen in table4, the collected database is more general than others.

## V. CONCLUSION

The main aim of this paper was to propose a robust algorithm to collect general databases which are used for image processing cases such as texture analysis and classification. In this respect, a feature representation is proposed based on combination of histogram statistical features, gray tone different matrixes and local binary patterns. Next, an innovative algorithm is prepared to make a texture database based on K-means clustering. In order to evaluate collection algorithm, the image candidates are represented using proposed feature extraction method, and were chosen as database member based on their distance.

The results are shown that the proposed feature representation provides good discrimination to analysis and classification textures. The results are shown that the database which is made based on proposed algorithm is more general than others and it covers texture analysis search space more than previous. Other advantages of this research are as follows:

1) Providing general and applicable texture database collecting algorithm which can be used for special cases.
2) Proposing feature representation method which can be used for very broad categories of image processing problems such as texture analysis, classification, image segmentation, visual defect detection and etc.
3) Low computational complexity of applying proposed feature representation against filter based and model based methods.
4) Solving the biasing to popular database problem which may be occurred for texture analysis researches.
5) Low noise sensitivity of proposed feature representation because of using global statistical features.

## VI. FUTURE WORK

The proposed collection algorithm is a general one which is not limited by a specific feature representation. In this respect, one interesting future research is to study for more accurate and general feature extraction methods to use. Collection databases which are included multi-sample classes, is a new problems in computer vision. The proposed collecting algorithm is not sensitive to the number of samples; therefore, it can be used for discussed problems as a future research.

## REFERENCES

[1] B.V. Reddy, M.R. Mani, B. Sujatha, V.V. Kumar- "Texture Classification Based on Random Threshold Vector Technique", International Journal of Multimedia and Ubiquitous Engineering, Vol. 5, No. 1, 2010, pp. 53-62

[2] Sh. Fekri-Ershad, "Color texture classification approach based on combination of primitive pattern units and statistical features, International Journal of Multimedia and its Applications, Vol. 3, No. 3, 2011, pp.1-13

[3] S. Liao, and A.C.S. Chung, Texture classification by using advanced local binary patterns and spatial distribution of dominant patterns, IEEE International Conference on Acoustics, Speech and Signal Processing, Vol. 1, 2007, pp. 1221-1224

[4] M.A. Akhloufi, X. Maldague, and W.B. Larbi, "A new color-texture approach for industrial products Inspection", Journal of multimedia, Vol. 3, No. 3, 2008, pp. 44-50

[5] X. Xie, "A review of recent advances in surface defect detection using texture analysis", Electronic Letters on Computer Vision and Image Analysis, Vol. 7, No. 3, 2008, pp. 1-22

[6] B.S. Anami, S.S. Nandyal, and A. Govardhan, "A combined color, texture and edge features based approach for identification and classification of Indian medical plants", International Journal of Computer Applications, Vol. 6, No. 12, 2010, pp. 45-51

[7] http://www.ux.uis.no/~tranden/brodatz.html

[8] http://www.outex.oulu.fi/

[9] T. Ojala, M. Pietikainen, and T. Maenpaa, "Multi resolution gray-scale and rotation invariant texture classification with local binary patterns", IEEE Transactions on Pattern Analysis and Machine Intelligence, Vol. 24, No. 7, 2002, pp. 971–987.

[10] R.M. Harlick, K. Shanmugam, and I. Dinstein, "Textural Features for Image Classification", IEEE Transactions on Systems and Cybernetics, Vol. 3, No. 6, 1973, pp. 610-621

[11] M. Berg, O. Cheong, M. van Kreveld, and M. Overmars, "Computational Geometry: Algorithms and Applications", Springer-Verlag, third edition, 2008.
[12] G. Castellano, L. Bonilha, L.M. Li, and F. Cendes, "Texture analysis of medical images", Clinical Radiology, Vol. 59, 2004, pp. 1061–1069
[13] A. Eleyan, and H. Demirel, "Co-occurrence matrix and its statistical features as a new approach for face recognition", Turkish Journal of Electronic Engineering& Computer Science, Vol.19, No.1, 2011, pp. 97-107
[14] A. Suresh, U.S.N. Raju and V. Vijaya Kumar, "An Innovative Technique of Stone Texture Classification Based on Primitive Pattern Units", International Journal of Signal and Image Processing, Vol.1, No. 1, 2010, pp. 40-45
[15] M. Amadasun, and R. King, "textural features corresponding to textural properties", IEEE Transactions on System, Man Cybernetics, Vol. 19, No. 5, 1989, pp. 1264-1274
[16] T. Maeenpaeae, M. Pietikaeinen, and T. Ojala, "Texture Classification by Multi Predicate Local Binary Pattern Operators", In Proc. of 15th International Conference on Pattern Recognition, vol. 3, 2000, pp. 951-954.
[17] M.Pietikäinen, T.Ojala, and Z.Xu, "Rotation-Invariant Texture Classification Using Feature Distributions", Pattern Recognition, Vol. 33, 2000, pp 43-52.
[18] F. Tajeripour and Sh. Fekri-Ershad, "Developing a Novel Approach for Stone Porosity Computing Using Modified Local Binary Patterns and Single Scale Retinex", Arabian Journal for Science and engineering, Vol. 39, No. 2, 2014, pp. 875-889
[19] Sh. Fekri-Ershad, and S. Hashemi, "To increase quality of feature reduction approaches based on processing input datasets", IEEE International Conference on Data Storage and Data Engineering, Vol. 1, 2011, pp. 367-372
[20] Sh. Fekri-Ershad, "Texture classification approach based on energy Variation", International Journal of Multimedia Technology, Vol. 2, No. 2, 2012, pp. 52-55
[21] S. Theodoridis, and K. Koutroumbas, "Pattern Recognition", Second Edition, Elsevier Academic Press, 2003
[22] V. Monev, "Introduction to similarity searching in chemistry", Match Communication in mathematical and in computer chemistry, Vol. 51, 2004, pp. 7-38
[23] S.H. Cha, "Taxonomy of nominal type histogram distance measures", In Proc. of American Conference on Applied Mathematics, Vol. 1, 2008, pp. 325-330